\documentclass[]{spie}  %>>> use for US letter paper
\usepackage{amsmath,amsfonts,amssymb,pifont,comment}
\usepackage{graphicx}
\usepackage[colorlinks=true, allcolors=blue]{hyperref}
\usepackage{multirow}
\usepackage[table]{xcolor}
\definecolor{mygray}{gray}{.9}
\usepackage{xcolor}

\usepackage{algorithm}
\usepackage{algpseudocode}

\title{Effective Dual-Region Augmentation for Reduced Reliance on Large Amounts of Labeled Data}

\author[a]{Prasanna Reddy Pulakurthi} 
\author[a]{Majid Rabbani}
\author[b]{Celso M. de Melo}
\author[a]{Sohail A. Dianat}
\author[b]{Raghuveer M. Rao}

\affil[a]{Rochester Institute of Technology, Rochester, NY, USA}
\affil[b]{DEVCOM Army Research Laboratory, Adelphi, MD, USA}

\begin{document} 
\maketitle

\begin{abstract}
This paper introduces a novel dual-region augmentation approach designed to reduce reliance on large-scale labeled datasets while improving model robustness and adaptability across diverse computer vision tasks, including source-free domain adaptation (SFDA) and person re-identification (ReID). Our method performs targeted data transformations by applying random noise perturbations to foreground objects and spatially shuffling background patches. This effectively increases the diversity of the training data, improving model robustness and generalization. Evaluations on the PACS dataset for SFDA demonstrate that our augmentation strategy consistently outperforms existing methods, achieving significant accuracy improvements in both single-target and multi-target adaptation settings. By augmenting training data through structured transformations, our method enables model generalization across domains, providing a scalable solution for reducing reliance on manually annotated datasets. Furthermore, experiments on Market-1501 and DukeMTMC-reID datasets validate the effectiveness of our approach for person ReID, surpassing traditional augmentation techniques. The code is available at \url{https://github.com/PrasannaPulakurthi/Foreground-Background-Augmentation}. 

\footnotetext{
\copyright\ 2025 Society of Photo-Optical Instrumentation Engineers (SPIE).  
This is the accepted version of the paper published in \textit{Proceedings of SPIE}, Vol. 13459,  
\textit{Synthetic Data for Artificial Intelligence and Machine Learning: Tools, Techniques, and Applications III}.  
The final published version is available at: \url{https://doi.org/10.1117/12.3058627}
}

\end{abstract}

\keywords{Data Augmentation, Classification, Source-Free Domain Adaptation, Person Re-Identification}

\section{INTRODUCTION}
\label{sec:intro}
% Introduce the problem statement. 
% 1. Deep learning models require a lot of data. 
% a. Therefore we developed a synthetic data generation method that helps build more robust models. Hence reducing the reliance on large labeled datasets. 
% b. In addition, the proposed generation method can be used as data augmentation to improve robustness and improve classification performance on the Person Re-ID task. 
% 3. And also can be used as an augmentation method for improving domain adaption. In this work, we particularly focus on source-free domain adaptation which is a more real work relevance and challenging paradigm because the source data is not available. 
% We demonstrate the effectiveness of our method by comparing it to other methods. In SFDA our data augmentation method added to Adacontrast improves performance by -% and -% in single and multi-target settings. 
% In-person re-id task improves the baseline performance by -% and -% R@1 score using the resent 18 and efficient net backbone. It also outperforms other augmentation methods in the literature. 
% In this work introduces the concept of applying different augmentations in the foreground and background. In the foreground, we randomly add noise patches, and in the background, we propose patch shuffling. 

In recent years, deep learning models have achieved remarkable performance across various computer vision tasks, including image classification, object detection, and person re-identification (ReID). However, achieving high accuracy typically requires extensive amounts of data, particularly labeled data, which are costly and labor-intensive to acquire. To address this challenge, we propose a novel dual-region augmentation method designed to reduce the dependence on extensive labeled datasets while enhancing model robustness and adaptability. 

\begin{figure}[ht]
    \begin{center}
    \resizebox{\columnwidth}{!}{
    \begin{tabular}{c} 
    \includegraphics[width=16.5cm]{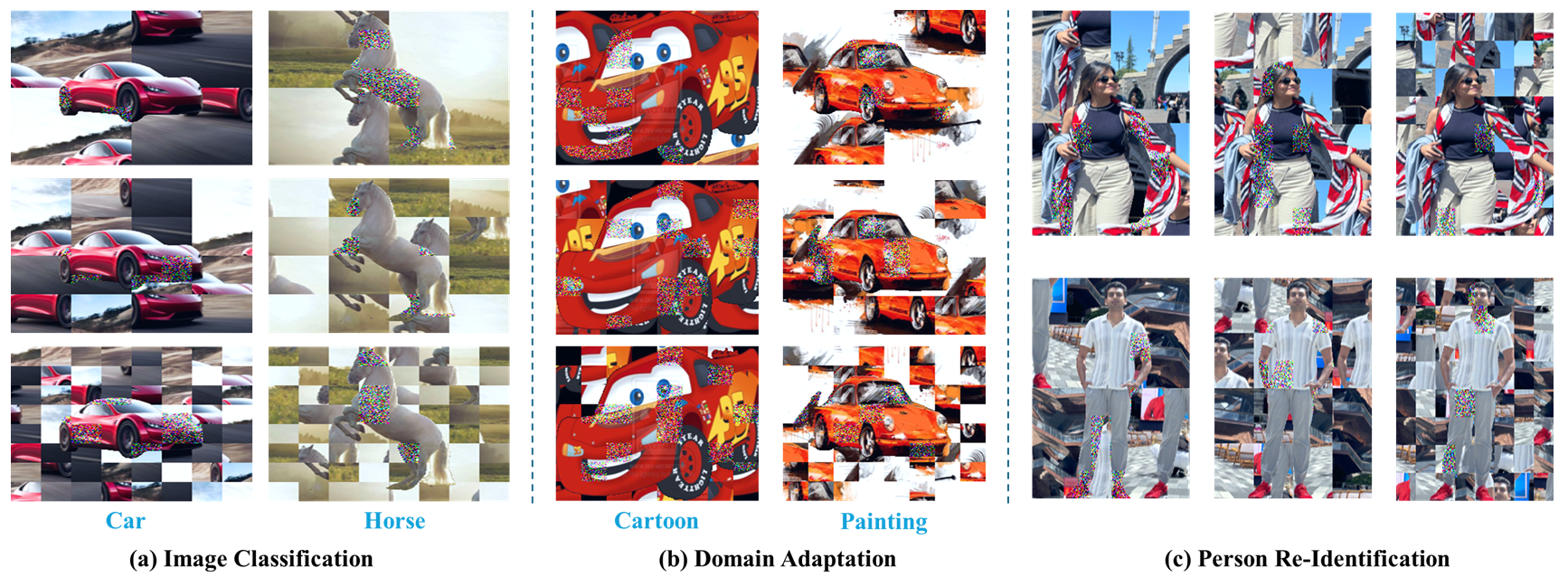} 
    \end{tabular}}
    \end{center}
    \caption[] 
    { \label{fig:examples} 
    Examples of the proposed augmentation method applied to (a) image classification, (b) domain adaptation, and (c) person re-identification, illustrating the integration of foreground patch noise and background patch shuffling.}
\end{figure}

Our approach introduces a dual-region data augmentation technique during training that applies distinct transformations to the foreground and background regions of images. Random noise patches are integrated into the foreground objects to introduce different levels of occlusion, while a patch-shuffling strategy is applied to the background to disrupt spatial consistency. This targeted transformation strategy increases the diversity of the training data, improving model robustness and generalization across varied conditions. Example images illustrating the proposed method are shown in Figure~\ref{fig:examples}.
 
Though deep learning models perform extremely well when training and testing data share the same distributions, their performance degrades significantly in the case of a domain shift~\cite{ds1}.
To address this issue, domain adaptation (DA) approaches have been proposed, leveraging the knowledge from a source domain to improve model performance on a related target domain. Among these techniques, unsupervised domain adaptation (UDA) has been a prominent technique, adapting models from a labeled source to an unlabeled target domain. Matching feature distributions between the target and source domains is one of the key strategies in UDA~\cite{dann,adda,cdan,famcd}. 
Source data access during adaptation is restricted in most real-world applications by privacy or logistical considerations.
This limitation has spawned source-free domain adaptation (SFDA), a specialized subcategory of UDA where a pre-trained source model must be adapted to a target domain without seeing the original source data. The absence of source data dramatically increases the complexity of adaptation, necessitating new methods to correctly realign the pre-trained source model to the target domain.

Our research specifically focuses on evaluating the dual-region augmentation approach in two distinct application scenarios: SFDA and person ReID. Our augmentation technique is integrated into leading SFDA methods, such as AdaContrast~\cite{adacontrast}, to demonstrate measurable performance improvements. Concurrently, we also evaluate our method on the person ReID task, where it consistently surpasses existing augmentation techniques, improving both model accuracy and generalization performance.
Overall, our work contributes a robust dual-region data augmentation strategy that effectively addresses the challenges posed by the limited availability of labeled data, enhancing performance in diverse scenarios such as classification, domain adaptation, and person ReID. 

Our key contributions are as follows:
\begin{enumerate}
    \item A dual region augmentation strategy is proposed, where Gaussian noise is applied to foreground regions and patch shuffling is performed on the background to promote diverse and robust feature learning.
    
    \item Evaluations are carried out on benchmark datasets for SFDA and person ReID, showing consistent improvements in generalization performance over strong baselines.
    
    \item An ablation study is provided to isolate the impact of each augmentation component, demonstrating that the combined approach yields the highest accuracy and robustness.
\end{enumerate}

In the next section, we review relevant literature and previous advancements in data augmentation and generation, particularly in the context of domain adaptation and person ReID. 

\section{Related Work}
\label{sec:related_work}
% 1. Talk about domain adaptation
% 2. One paragraph on UDA
% 3. SFDA
% 4. Augmentation Methods
% 5. Person Re-identification

\subsection{Data Augmentation and Generation in Domain Adaptation}
Data augmentation and generation techniques are pivotal in domain adaptation, enhancing model robustness across varying data distributions. Generative Adversarial Networks (GANs) \cite{goodfellow2014generative} are often utilized to synthesize realistic images~\cite{Karras2020ada, 10732016}, thereby bridging the gap between source and target domains \cite{pmlr-v80-hoffman18a, liu2019liquid, pulakurthi2024enhancing}. In addition, AugGAN~\cite{huang2018auggan} utilizes GAN-based data augmentation to facilitate cross-domain adaptation by preserving image-object integrity and maintaining translation consistency, thereby generating more visually plausible images across different domains. LearnAug-UDA~\cite{Carrazco_2023_BMVC} employs an encoder-decoder architecture that transforms source data to resemble the target domain, thereby improving classifier generalization.
Adversarial and Random Transformations~\cite{xiao2023adversarial} combine adversarial and random transformations to enhance model robustness in domain adaptation and generalization tasks. These methodologies collectively underscore the significance of data augmentation and generation in mitigating domain shifts and enabling models to generalize effectively across diverse domains.

\subsection{Data Augmentation and Generation in Person Re-Identification}
Data augmentation and generation techniques play a crucial role in improving the robustness and accuracy of person ReID systems. Several methods have been developed to generate synthetic data, enhancing Person ReID performance by addressing challenges such as occlusions, pose variations, and domain shifts~\cite{zheng2022person, zheng2019joint, zheng2017unlabeled}. One such approach, Improved CycleGAN~\cite{YANG2023100409}, extends CycleGAN~\cite{zhu2017unpaired} to generate synthetic images tailored for Person ReID, effectively mitigating issues related to pose variability and occlusions. Additionally, widely used augmentation techniques, such as Random Erasing~\cite{zhong2020random}, randomly remove regions of an image during training to simulate occlusions, thereby improving model resilience. Furthermore, grayscale transformation techniques, including Random Grayscale Transformation and Random Grayscale Patch Replacement~\cite{gong2021person}, have been employed to enhance model robustness against illumination changes. These augmentation and generation strategies contribute to improving generalization and performance in person ReID tasks.

\section{Methodology}
\label{sec:method}
\begin{figure}[t]
    \begin{center}
    \resizebox{\columnwidth}{!}{
    \begin{tabular}{c} 
    \includegraphics[width=16cm]{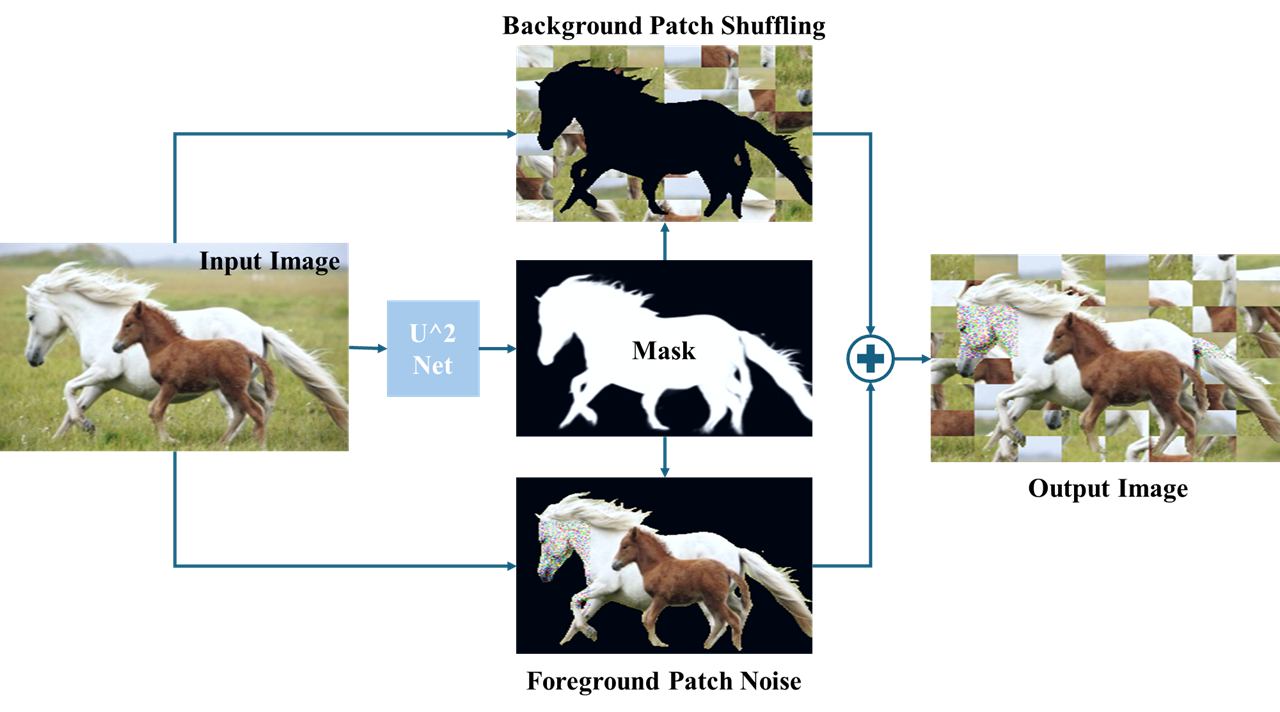} 
    \end{tabular}}
    \end{center}
    \caption[] 
    { \label{fig:method} 
    Overview of the proposed data augmentation pipeline. Foreground and background regions are separated using a U$^2$-Net~\cite{Qin_2020_PR} generated mask, followed by distinct augmentations: Gaussian noise applied to the foreground and patch shuffling applied to the background. The modified regions are then recombined to form the augmented image.}
\end{figure}

This section describes the proposed data augmentation method, which applies separate transformations to the foreground and background regions of input images. The overall pipeline is illustrated in Figure~\ref{fig:method} and the pseudo-code is presented in Algorithm \ref{algo:FBA}. The goal of this augmentation is to improve model robustness by introducing structured perturbations that encourage holistic feature learning and reduce reliance on background-specific cues.

\textbf{Foreground-Background Segmentation:}
To augment the foreground and background separately, we first generate a segmentation mask using the U$^2$-Net~\cite{Qin_2020_PR} model. U$^2$-Net is a deep convolutional neural network designed for salient object detection, recognized for its high accuracy. The model outputs a probability mask, which is converted into a binary mask via thresholding. This mask isolates the foreground object from the background, enabling targeted augmentation.

\begin{algorithm}[t]
\caption{Foreground-Background Augmentation}\label{algo:FBA}
\textbf{Input:} Input image $I$; Segmentation Mask $M$; Image size $W \times H$; Mixing probability $\rho$; \\
Patch size set $\mathcal{P}$, For example, if $W=H=224$, then $\mathcal{P} = \{14, 28, 56, 112\}$; Noise area ratio range $[a_l, a_h]$; \\
\textbf{Output:} Augmented image $I^*$\\
\textbf{Initialization:} $p_1 \leftarrow \text{Rand}(0, 1)$

\begin{algorithmic}[1]
\If{$p_1 \geq \rho$}
    \State $I^* \leftarrow I$
    \State \textbf{return} $I^*$
\Else
    \Statex \Comment{Foreground Patch Noise}
    \State $A \leftarrow \mathcal{U}(a_l, a_h) \; \text{Area}(I)$
    \State Select $k$ random patches $\{P_1, \dots, P_k\}$ such that $\sum \text{Area}(P_i) \approx A$
    \ForAll{patch $P_i$}
        \State Apply Gaussian noise to pixels in $P_i$
    \EndFor
    \State $I_{fg} \leftarrow I$ with noise applied to selected patches

    \Statex \Comment{Background Patch Shuffling}
    \State $(P_h, P_w) \leftarrow \text{RandomChoice}(\mathcal{P})$
    \State Divide $I$ into $N = \left\lfloor \frac{W}{P_w} \right\rfloor \times \left\lfloor \frac{H}{P_h} \right\rfloor$ patches
    \State Shuffle the $N$ patches randomly
    \State $I_{bg} \leftarrow$ Reconstruct image from shuffled patches

    \Statex \Comment{Restore Original Foreground}
    \ForAll{pixels $(i, j)$ in $I$}
        \If{$M(i, j) = 1$}
            \State $I^*(i, j) \leftarrow I_{fg}(i, j)$
        \Else
            \State $I^*(i, j) \leftarrow I_{bg}(i, j)$
        \EndIf
    \EndFor
    \State \textbf{return} $I^*$
\EndIf
\end{algorithmic}
\end{algorithm}

\textbf{Dual-Region Data Augmentation:}
Once the mask is obtained, the augmentation process follows two distinct paths:

\textbf{1. Foreground Patch Noise (FPN):}
The augmentation starts by applying Gaussian noise patches to randomly selected regions across the entire image, introducing noisy perturbations. The total area covered by these noise patches is randomly determined by sampling from a uniform distribution $\mathcal{U}(a_l, a_h)$ between $a_l=2\%$ and $a_h=40\%$ of the full image area. Subsequently, the foreground object is extracted using the binary segmentation mask, resulting in a foreground region that contains the introduced noise. 
The primary objective of this augmentation is to improve the model's robustness by introducing localized perturbations, which encourage it to utilize a broader set of discriminative features rather than relying on specific regions. For instance, without augmentation, a network may disproportionately focus on the horse’s head for classification. By applying noise distortions, the model is compelled to recognize a wider range of distinguishing characteristics, ultimately enhancing its ability to generalize across different variations of the object.

\textbf{2. Background Patch Shuffling (BPS):}
The entire image is first divided into a grid of randomly chosen dimensions, resulting in a total of 2×2, 4×4, or 8×8 patches. All patches, including both foreground and background, are then randomly shuffled to disrupt spatial coherence. After shuffling, the original foreground region is restored using the segmentation mask, ensuring that only the background remains perturbed. This augmentation prevents the model from overfitting to background-specific cues, forcing it to focus on the foreground object for classification. By randomly varying patch sizes across images, the method increases dataset diversity and enhances model generalization. 
Since patches are shuffled before the foreground is restored, fragments of the foreground may still appear in the background. This encourages the model to learn object-level representations rather than focusing on isolated discriminative regions, promoting holistic feature learning.

The augmented foreground with noise patches and the shuffled background are recombined to form a composite output image with distinctly augmented foreground and background regions. This augmentation strategy is applied to both SFDA and person ReID tasks, where experiments demonstrate its effectiveness in improving model performance and highlight its versatility and applicability in different computer vision tasks. The next section outlines the experimental setup used to evaluate the proposed method.

\section{Experimental Setup}
\label{sec:experiments}

\subsection{Source-Free Domain Adaptation}
\textbf{Dataset:}
The PACS dataset~\cite{pacs} is used to evaluate the proposed method. It contains four distinct visual domains: Photo (1,670 images), Art Painting (2,048 images), Cartoon (2,344 images), and Sketch (3,929 images). These domains consist of seven object categories with considerable domain shifts across different styles. 

We follow the evaluation protocols described in NEL~\cite{nel}, measuring and comparing average accuracy across six defined source-target domain combinations for both single-target and multi-target adaptation settings.
As the AdaContrast~\cite{adacontrast} baseline did not originally report results on PACS, we generated these baseline results using their publicly available codebase, after verifying consistency with their published benchmarks. Our reproduced AdaContrast results (presented in Tables \ref{tab:pacs_single_target} and \ref{tab:pacs_multi_target}) surpass previously reported best results in NEL. Our proposed method is integrated as an additional augmentation strategy within the AdaContrast framework.

\textbf{Backbone:}
We adopt ResNet-18~\cite{he2016deep} as our feature-extraction backbone, following the standard practice for domain adaptation experiments on PACS.

\textbf{Hyperparameters:}
Experiments utilize Stochastic Gradient Descent (SGD) with a learning rate of 2e-4, a batch size of 128, three nearest neighbors, and 100 epochs. The proposed augmentation is applied to a fraction ($\rho$) of the entire set of augmentations: $\rho = 0.8$ in single-target settings and $\rho = 0.5$ in multi-target settings.

\subsection{Person Re-Identification}
\textbf{Datasets:} 
We evaluate the proposed method against existing augmentation techniques using two standard benchmarks: Market-1501~\cite{Zheng_2015_ICCV} and DukeMTMC-reID~\cite{ristani2016performance}. 
The Market-1501 dataset comprises 32,668 automatically detected pedestrian bounding-box images of 1,501 identities, captured across six surveillance cameras. It is partitioned into a training set (12,936 images, 751 identities) and a testing set (19,732 images, 750 identities), with an additional 3,368 query images used for evaluation. 
The DukeMTMC-reID dataset consists of manually annotated pedestrian images extracted from videos recorded by eight surveillance cameras. It includes 16,522 training images (702 identities), 17,661 gallery images, and 2,228 query images (702 different identities). The query images are selected as a probe set to find the correct match across reference gallery images.

\textbf{Backbone:} We conduct experiments using two widely adopted convolutional neural networks: ResNet-18~\cite{he2016deep} and EfficientNet-B4~\cite{tan2019efficientnet}.

\textbf{Hyperparameters:} 
All person ReID experiments utilize the Stochastic Gradient Descent (SGD) optimizer with a momentum of 0.9, an initial learning rate of 0.05, a weight decay of $5 \times 10^{-4}$, mixing probability $\rho = 0.5$, and a batch size of 32. Each model is trained for 30 epochs.

\section{Results}
\label{sec:results}

\subsection{Source-Free Domain Adaptation}
We evaluated the effectiveness of our proposed augmentation method on the PACS~\cite{pacs} dataset using ResNet-18~\cite{he2016deep} as the backbone. The results for both single-target and multi-target SFDA settings are summarized in Tables~\ref{tab:pacs_single_target} and~\ref{tab:pacs_multi_target}.
In the single-target SFDA setting (Table~\ref{tab:pacs_single_target}), our method achieves the highest overall average accuracy of \textbf{84.0\%}, surpassing both AdaContrast~\cite{adacontrast} (79.4\%) and NEL~\cite{nel} (72.4\%). Notably, our approach demonstrates substantial improvements in challenging adaptation tasks such as Photo to Sketch (P→S), improving accuracy from 66.7\% (AdaContrast) to \textbf{77.0\%}, and Art-Painting to Sketch (A→S), increasing from 77.9\% to \textbf{85.3\%}.
In the multi-target SFDA setting (Table~\ref{tab:pacs_multi_target}), our proposed method consistently outperforms existing methods, achieving an average accuracy of \textbf{77.4\%}, compared to AdaContrast’s 75.4\% and NEL’s 68.4\%. Our approach significantly enhances performance on challenging domains like Sketch, with improvements from 62.9\% (AdaContrast) to \textbf{65.6\%} for P→S, and from 72.9\% to \textbf{77.7\%} for A→S.
These results demonstrate that our augmentation approach greatly enhances model robustness and generalization for domain adaptation tasks. By effectively enhancing performance on unlabeled target domains, our approach can mitigate the reliance on target labels.

%%%%%%%%%%%%%%%%%%%%%%%%%%%%%%%%%%%%%%%%%%%%%%%%%%%%%%%%%%%%%%%%%%%%%%%%%%%%%%%%
% PACS
\begin{table}[t]
\caption{Classification accuracy (\%) on PACS~\cite{pacs} for the single-target setting with ResNet-18. Legend: \textbf{P}: Photo, \textbf{A}: Art-Painting, \textbf{C}: Cartoon, and \textbf{S}: Sketch. The highest accuracies are in \textbf{bold}. * indicates our reproduced results using \cite{adacontrast}.}
\label{tab:pacs_single_target}
\begin{center}
\renewcommand{\arraystretch}{1.3}
\footnotesize
\begin{tabular}{l | c |c c c c c c|c}
\hline

\hline
\rowcolor{mygray}
\textbf{Method} & \textbf{Source-Free} & \textbf{P→A} & \textbf{P→C} & \textbf{P→S} & \textbf{A→P} & \textbf{A→C} & \textbf{A→S} & \textbf{Average} \\
\hline
\hline
% Source & 60.6 & 22.6 & 22.4 & 96.0 & 49.9 & 37.3 & 48.1 \\
% Target & 93.7 & 91.5 & 94.7 & 99.4 & 91.5 & 94.5 & 94.2 \\
% \hline
NEL \cite{nel} & \ding{51} & 82.6 & \textbf{80.5} & 32.3 & 98.4 & \textbf{84.3} & 56.1 & 72.4 \\
AdaContrast \cite{adacontrast}* & \ding{51}  & 81.3 & 72.2 & 66.7 & \textbf{98.7} & 79.7 & 77.9 & 79.4 \\
\hline
% Background PatchShuffle (Ours) & 83.5 &	75.9 &	71.4 &	\textbf{99.0} &	82.3 &	78.9 &	81.8 \\

% Ours + RE & 82.4 &	78.6 &	76.0 &	98.9 &	82.9 &	78.4 &	81.8 \\

% Ours + RPN (Ours) & \textbf{84.9} & 77.9 &	72.8 &	98.7 &	83.0 &	79.2 &	82.8 \\

Ours & \ding{51} & \textbf{83.7} & 76.4 & \textbf{77.0} & 98.4 & 82.9 & \textbf{85.3} & \textbf{84.0} \\

\hline

\hline
\end{tabular}
\end{center}
\end{table}

%%%%%%%%%%%%%%%%%%%%%%%%%%%%%%%%%%%%%%%%%%%%%%%%%%%%%%%%%%%%%%%%%%%%%%%%%%%%%%%%
% PACS Multi-target
\begin{table}[t]
\caption{Classification accuracy (\%) on PACS~\cite{pacs} for the multi-target setting. All the methods use the ResNet-18 backbone. Legend: \textbf{P}: Photo, \textbf{A}: Art-Painting, \textbf{C}: Cartoon, and \textbf{S}: Sketch. The highest value is \textbf{bolded} and * indicates our reproduced results on the AdaContrast~\cite{adacontrast} baseline .}
\label{tab:pacs_multi_target}
\begin{center}
\renewcommand{\arraystretch}{1.3}
\footnotesize
\begin{tabular}{l|c|c c c|c c c|c}
\hline

\hline
\rowcolor{mygray}
\multicolumn{2}{c|}{\textbf{Multi-Target UDA}} & \multicolumn{3}{c|}{\textbf{P → ACS}} & \multicolumn{3}{c|}{\textbf{A → PCS}} & \\
\hline
\rowcolor{mygray}
\textbf{Method} & \textbf{Source-Free} & \textbf{A} & \textbf{C} & \textbf{S} & \textbf{P} & \textbf{C} & \textbf{S} & \textbf{Average} \\
\hline
\hline
% 1-NN & \ding{55} & 15.2 & 18.1 & 25.6 & 22.7 & 19.7 & 22.7 & 20.7 \\
ADDA \cite{adda} & \ding{55} & 24.3 & 20.1 & 22.4 & 32.5 & 17.6 & 18.9 & 22.6 \\
DSN \cite{dsn} & \ding{55} & 28.4 & 21.1 & 25.6 & 29.5 & 25.8 & 26.8 & 25.8 \\
ITA \cite{ita} & \ding{55} & 31.4 & 23.0 & 28.2 & 35.7 & 27.0 & 28.9 & 29.0 \\
KD \cite{kd} & \ding{55} & 24.6 & 32.2 & 33.8 & 35.6 & 46.6 & 57.5 & 46.6 \\
NEL \cite{nel} & \ding{51} & \textbf{80.1} & 76.1 & 25.9 & \textbf{96.0} & \textbf{82.8} & 49.8 & 68.4 \\
AdaContrast \cite{adacontrast}* & \ding{51} & 70.1 & 77.9 & 62.9 & 95.9 & 72.7 & 72.9 & 75.4 \\
\hline
% Ours & \checkmark & 71.9 &	74.7 &	\textbf{65.5} &	94.5 &	71.0 &	\textbf{77.3} &	\textbf{75.9} \\
% Ours & \ding{51} & 70.4 &	\textbf{78.1} &	\textbf{63.2} &	95.2 &	71.8 & \textbf{69.9} &	\textbf{75.9} \\
% Ours & \ding{51} & 73.0 &	\textbf{78.9} &	\textbf{63.3} &	95.0 &	70.4 & \textbf{73.9} &	\textbf{75.8} \\
Ours & \ding{51} & 73.2 &	\textbf{78.5} &	\textbf{65.6} &	95.9 &	73.5 & \textbf{77.7} &	\textbf{77.4} \\
\hline

\hline
\end{tabular}
\end{center}
\end{table}

%%%%%%%%%%%%%%%%%%%%%%%%%
\textbf{Ablation Study:}
An ablation study was conducted to systematically evaluate the contributions of individual augmentation techniques, specifically Background Patch Shuffle (BPS) and Foreground Patch Noise (FPN), to overall model performance (Table \ref{tab:ablation}). Starting from the MoCo v2~\cite{chen2020improved} baseline augmentation used in AdaContrast, each augmentation method was added to the baseline to assess its impact.

Incorporating Background Patch Shuffle increased average accuracy from 79.4\% to 82.2\%, with notable improvements in challenging adaptations such as Photo to Sketch (P→S), where accuracy rose from 66.7\% to 71.3\%. Independently, adding Foreground Patch Noise significantly enhanced performance, boosting accuracy to 83.6\%, particularly in scenarios like P→S and Art-Painting to Cartoon (A→C). Combining both augmentations resulted in the highest average accuracy of 84.0\%, highlighting the complementary effectiveness of foreground noise integration and background shuffling. These findings indicate that the proposed combined augmentation strategy noticeably improves robustness and generalization across diverse domain adaptation tasks.

%%%%%%%%%%%%%%%%%%%%%%%%%%%%%%%%%%%%%%%%%%%%%%%%%%%%%%%%%%%%%%%%%%%%%%%%
% Ablation
\begin{table}[t]
\caption{Ablation study evaluating the effect of individual augmentations on classification accuracy (\%) for the PACS~\cite{pacs} dataset in the single-target setting using ResNet-18. Legend: \textbf{P}: Photo, \textbf{A}: Art-Painting, \textbf{C}: Cartoon, and \textbf{S}: Sketch.}
\label{tab:ablation}
\begin{center}
\renewcommand{\arraystretch}{1.3}
\small
\resizebox{\columnwidth}{!}{
\begin{tabular}{c c c|c c c c c c|c}
\hline

\hline
\rowcolor{mygray}
% \multicolumn{3}{c|}{\textbf{Augmentations}} & \multicolumn{6}{c|}{\textbf{Domain Shift}} & \\
% \hline
% Source & 60.6 & 22.6 & 22.4 & 96.0 & 49.9 & 37.3 & 48.1 \\
% Target & 93.7 & 91.5 & 94.7 & 99.4 & 91.5 & 94.5 & 94.2 \\
% \hline
\rowcolor{mygray}
\textbf{MoCo v2}~\cite{chen2020improved} & \textbf{Background} & \textbf{Foreground} & & & & & & &\\
\rowcolor{mygray}
\textbf{(AdaContrast}~\cite{adacontrast}\textbf{)} & \textbf{Patch Shuffle} & \textbf{Patch Noise} &  \multirow{-2}{*}{\textbf{P→A}} & \multirow{-2}{*}{\textbf{P→C}} &  \multirow{-2}{*}{\textbf{P→S}} &  \multirow{-2}{*}{\textbf{A→P}} &  \multirow{-2}{*}{\textbf{A→C}} &  \multirow{-2}{*}{\textbf{A→S}} &  \multirow{-2}{*}{\textbf{Average}} \\
\hline
\hline
\ding{51} & \ding{55} & \ding{55}  & 81.3 & 72.2 & 66.7 & 98.7 & 79.7 & 77.9 & 79.4 \\
% \ding{51} & \ding{51} & \ding{55} & 83.5 &	75.9 &	71.4 &	\textbf{99.0} &	82.3 &	78.9 &	81.8 \\
% \ding{51} & \ding{51} & \ding{51} & \ding{55} & 82.4 &	78.6 &	76.0 &	98.9 &	82.9 &	78.4 &	81.8 \\
% \ding{51} & \ding{55} & \ding{51} & \textbf{84.9} & 77.9 &	72.8 &	98.7 &	\textbf{83.0} &	79.2 &	82.8 \\
% \ding{51} & \ding{51} & \ding{51} & 84.1 & 79.2 & 75.8 &	98.7 &	82.8 &	79.5 &	\textbf{83.4} \\
\ding{51} & \ding{51} & \ding{55} & 83.5 & 72.9 & 71.3 & 98.9 &	82.2 & 84.6 & 82.2 \\
\ding{51} & \ding{55} & \ding{51} & 84.1 & 78.6 & 74.1 & 99.0 & 83.7 & 82.2 & 83.6 \\
\ding{51} & \ding{51} & \ding{51} & 83.7 & 76.4 & 77.0 & 98.4 & 82.9 & 85.3 & 84.0 \\
\hline

\hline
\end{tabular}}
\end{center}
\end{table}

%%%%%%%%%%%%%%%%%%%%%%%%%%%%%%%%%%%%%%%%%%%%%%%%%%%%%%%%%%%%%%%%%%%%%%%%
\subsection{Person Re-Identification} We evaluated our augmentation method on Market-1501~\cite{Zheng_2015_ICCV} and DukeMTMC-reID~\cite{ristani2016performance} datasets using ResNet-18~\cite{he2016deep} and EfficientNet-b4~\cite{tan2019efficientnet} backbones (Table~\ref{tab:personreid}). Our method consistently outperformed both the baseline and Random Erasing~\cite{zhong2020random}. Notably, the most significant gains were observed with the EfficientNet-b4 backbone on the DukeMTMC-reID dataset, where Rank@1 accuracy increased substantially from 74.60\% (baseline) and 80.16\% (Random Erasing) to \textbf{81.51\%}. Similarly, mean Average Precision (mAP) significantly improved from 54.91\% (baseline) and 62.26\% (Random Erasing) to \textbf{64.07\%}. Comparable improvements were observed on Market-1501, with Rank@1 accuracy increasing from 82.99\% (baseline) and 87.20\% (Random Erasing) to \textbf{87.41\%}, and mAP rising from 62.05\% (baseline) and 68.71\% (Random Erasing) to \textbf{69.85\%}. These results validate the effectiveness of our augmentation technique, demonstrating substantial improvements in both robustness and generalization performance in the person re-identification task.

%%%%%%%%%%%%%%%%%%%%%%%%
% Person Re-ID
\begin{table}[t]
    \caption{Comparison of person re-identification performance across different backbones and augmentation strategies on Market-1501~\cite{Zheng_2015_ICCV} and DukeMTMC-reID~\cite{ristani2016performance} datasets. The table reports Rank@1, Rank@5, Rank@10, and mean Average Precision (mAP) for each method.}
    
    \begin{center}
    \label{tab:personreid}
    \renewcommand{\arraystretch}{1.3}
    \small
    \resizebox{\columnwidth}{!}{
    \begin{tabular}{l|c|l|cccc|cccc}
    \hline
    
    \hline
    \rowcolor{mygray}
    & & & \multicolumn{4}{c|}{\textbf{DukeMTMC-reID}} & \multicolumn{4}{c}{\textbf{Market-1501}} \\
    \cline{4-11}
    \rowcolor{mygray}
    \multirow{-2}{*}{\textbf{Backbone}} & \multirow{-2}{*}{\textbf{Parameters}} & \multirow{-2}{*}{\textbf{Augmentation}} & \textbf{R@1$\uparrow$} & \textbf{R@5}$\uparrow$ & \textbf{R@10}$\uparrow$ & \textbf{mAP}$\uparrow$ & \textbf{R@1}$\uparrow$ & \textbf{R@5}$\uparrow$ & \textbf{R@10}$\uparrow$ & \textbf{mAP}$\uparrow$ \\
    \hline
    \hline
    % ----------------- ResNet-18 ------------------
    % Rank@1:0.633752 Rank@5:0.780072 Rank@10:0.828097 mAP:0.401748
    % Rank@1:0.668646 Rank@5:0.844715 Rank@10:0.891924 mAP:0.394383
    \multirow{4}{*}{ResNet-18~\cite{he2016deep}} & \multirow{4}{*}{11.8 M} & 
    Baseline & 63.38 & 78.01 & 82.81 & 40.18 & 66.86 & 84.47 & 89.19 & 39.43 \\
    % Rank@1:0.611759 Rank@5:0.773339 Rank@10:0.827199 mAP:0.395926
    % Rank@1:0.681710 Rank@5:0.854513 Rank@10:0.900238 mAP:0.410180
    & & Random Grayscale~\cite{gong2021person} & 61.18 & 77.33 & 82.72 & 39.59 & 68.17 & 85.45 & 90.02 & 41.02 \\
    % Rank@1:0.646768 Rank@5:0.802065 Rank@10:0.846499 mAP:0.439517
    % Rank@1:0.711105 Rank@5:0.872328 Rank@10:0.920131 mAP:0.454128
    & & Random Erasing~\cite{zhong2020random} & 64.68 & 80.21 & 84.65 & 43.95 & 71.11 & 87.23 & 92.01 & 45.41 \\
    % Rank@1:0.678636 Rank@5:0.815978 Rank@10:0.859066 mAP:0.458490
    % Rank@1:0.713777 Rank@5:0.871140 Rank@10:0.919537 mAP:0.469497
    \cline{3-11}
    & & Ours & \textbf{67.86} & \textbf{81.60} & \textbf{85.91} & \textbf{45.85} & \textbf{71.38} & \textbf{87.11} & \textbf{91.95} & \textbf{46.95} \\
    \hline
    \hline
    % ----------------- EfficientNet-b4 ------------------
    \multirow{4}{*}{EfficientNet-b4~\cite{tan2019efficientnet}} & \multirow{4}{*}{20.6 M} & 
    % Rank@1:0.745960 Rank@5:0.852783 Rank@10:0.897217 mAP:0.549123
    % Rank@1:0.829869 Rank@5:0.933492 Rank@10:0.959620 mAP:0.620462
    Baseline & 74.60 & 85.28 & 89.72 & 54.91 & 82.99 & 93.35 & 95.96 & 62.05 \\
    % Rank@1:0.761670 Rank@5:0.872531 Rank@10:0.902603 mAP:0.556813
    % Rank@1:0.836995 Rank@5:0.938539 Rank@10:0.961401 mAP:0.622822
    & & Random Grayscale~\cite{gong2021person} & 76.17 & 87.25 & 90.26 & 55.68 & 83.70 & 93.85 & 96.14 & 62.28 \\
    % Rank@1:0.801616 Rank@5:0.898564 Rank@10:0.922352 mAP:0.622559
    % Rank@1:0.872031 Rank@5:0.951306 Rank@10:0.966449 mAP:0.687147
    & & Random Erasing~\cite{zhong2020random} & 80.16 & 89.86 & 92.24 & 62.26 & 87.20 & 95.13 & 96.64 & 68.71 \\
    \cline{3-11}
    % Rank@1:0.815081 Rank@5:0.902154 Rank@10:0.928187 mAP:0.640662
    % Rank@1:0.874109 Rank@5:0.952197 Rank@10:0.970606 mAP:0.698476
    & & Ours & \textbf{81.51} & \textbf{90.21} & \textbf{92.82} & \textbf{64.07} & \textbf{87.41} & \textbf{95.22} & \textbf{97.06} & \textbf{69.85} \\
    \hline

    \hline
    \end{tabular}}
    \end{center}
\end{table}

\section{Conclusion}
\label{sec:conclusion}
In this work, we introduced a novel dual-region data augmentation technique aimed at reducing reliance on large labeled datasets by leveraging targeted augmentation strategies. 
Specifically, our approach applies foreground noise integration and background patch shuffling. By separately augmenting these regions, our method encourages robust feature learning, significantly improving model generalization across diverse scenarios. This augmentation strategy effectively minimizes the need for extensive manual annotations.
Experiments across source-free domain adaptation (SFDA) and person re-identification (ReID) tasks validate the effectiveness of our method. Our method surpassed prior methods, significantly improving both single-target (84.0\%) and multi-target (77.4\%) adaptation accuracy. Additionally, our method outperformed baseline augmentations on standard person ReID benchmarks, yielding considerable Rank@1 and mean Average Precision (mAP) improvements on Market-1501 and DukeMTMC-reID.
By demonstrating that targeted data augmentations can enhance model robustness across different tasks, this work highlights a promising augmentation-based direction for reducing dependency on large-scale labeled data. 
Future work may refine these augmentation strategies, broaden their applicability to more diverse datasets, and evaluate their impact on other real-world computer vision applications.

\acknowledgments % equivalent to \section*{ACKNOWLEDGMENTS}       
This work was supported by DEVCOM U.S. Army Research Laboratory through Booz Allen Hamilton under contract W911QX-21-D-0001.

% References
\bibliography{report} % bibliography data in report.bib
\bibliographystyle{spiebib} % makes bibtex use spiebib.bst

\end{document}